\documentclass[9pt,twocolumn,twoside]{pnas-new}

\providecommand{\articletype}[1]{}
\providecommand{\Firstpage}{}
\providecommand{\Endparasplit}{}
\providecommand{\dataavail}[1]{%
  \section*{Data Availability}
  \small #1}
\providecommand{\bibsplit}[1][]{}

\articletype{Computer Sciences}
\templatetype{pnasresearcharticle}

\usepackage{amsmath,amssymb}
\usepackage{booktabs}
\usepackage{graphicx}
\usepackage{url}
\begin{document}

\title{DiRe -- RAPIDS: Topology-faithful dimensionality reduction at scale}

\author[a]{Alexander Kolpakov}
\author[b,1]{Igor Rivin}

\affil[a]{CSTEM, University of Austin, Austin, TX, USA}
\affil[b]{Department of Mathematics, Temple University, Philadelphia, PA, USA}

\leadauthor{Kolpakov}

\significancestatement{Dimensionality reduction methods such as UMAP and
t-SNE are central tools for visualising high-dimensional data, but their
local-neighborhood objectives can preserve sampling noise while distorting
global topology. We show that standard local metrics reward this noise
memorisation: top-performing embeddings invent cycles and disconnected
islands absent from the data. We introduce a topology-faithfulness benchmark
based on noisy manifolds with known homology, tune DiRe against it, and find
Pareto-optimal configurations that match or beat GPU-accelerated UMAP on
classification while recovering exact first Betti numbers on stress tests.
On 723K arXiv paper embeddings, DiRe preserves 3--4$\times$ more topological
structure than UMAP at comparable wall-clock.}

\authorcontributions{A.K. and I.R. designed methods, performed research,
contributed new analytic tools, analyzed data, and wrote the manuscript.}
\authordeclaration{The authors declare no competing interests.}
\correspondingauthor{\textsuperscript{1}To whom correspondence should be
addressed. E-mail: rivin@temple.edu}


\begin{abstract}
We revisit the evaluation of dimensionality-reduction algorithms in light of
the observation that metrics based on $k$-nearest-neighbor preservation reward
faithful reproduction of sampling noise as well as manifold structure. We
propose a direct replacement: \emph{topology error} on noisy-manifold stress
tests, measuring deviation from known first Betti numbers in the 2-D embedding
via \texttt{ripser}-computed persistence diagrams. Using this metric alongside
classification accuracy as objectives in a multi-objective NSGA-II search, we
tune the DiRe force-directed embedding method and compare against
GPU-accelerated cuML UMAP on 11 OpenML datasets and on a
723-thousand-paper arXiv embedding corpus. On 7 of 11 OpenML datasets, tuned
DiRe strictly Pareto-dominates cuML UMAP on both axes; on \emph{every}
dataset, a DiRe configuration exists that recovers exact first Betti numbers
on both stress manifolds while UMAP never drops below a topology error of 5.
On the arXiv corpus, DiRe preserves 3--4$\times$ more topological structure
(Betti-curve DTW distance) than UMAP at comparable wall-clock (20\,s vs 32\,s
for 723K points on a single H100-class GPU). The resulting hyperparameter
preset is published as part of the open-source DiRe library.
\end{abstract}

\dates{This manuscript was compiled on \today}
\doi{\url{www.pnas.org/cgi/doi/10.1073/pnas.XXXXXXXXXX}}

\maketitle
\thispagestyle{firststyle}
\ifthenelse{\boolean{shortarticle}}{\ifthenelse{\boolean{singlecolumn}}{\abscontentformatted}{\abscontent}}{}

\Firstpage
\label{sec:intro}

Dimensionality reduction methods such as t-SNE \citep{tsne} and UMAP
\citep{umap} are the standard tools for two-dimensional visualisation of
high-dimensional datasets, from single-cell expression profiles to language
model embeddings. Both methods optimise a local-neighborhood-preservation
objective: the embedding is good to the extent that each point's $k$
nearest neighbors in the original space remain its $k$ nearest neighbors
in the embedding. This design emphasises cluster structure and short-range
relationships, and both methods are effective at it.

The local objective is, however, known to come at the expense of global
and topological structure. UMAP in particular produces embeddings whose
2-D layouts introduce gaps, folds, and spurious cycles that are not in
the underlying data \citep{pachter,chari-pachter-2023}. These artefacts
are visually striking and scientifically misleading: two papers appearing
in separate UMAP islands need not be more dissimilar than two papers in
the same island, yet published biological UMAPs are routinely read as
though they were.

We make three contributions. First, we argue that the local-neighborhood
metric rewards reproduction of sampling noise as well as manifold
structure. We demonstrate this by sweeping noise level on manifolds with
known topology and showing that UMAP's $\beta_1$ estimate tracks the
inflating noisy-sample $\beta_1$, while a force-directed alternative
(DiRe, \citep{dire-joss}) stays near the theoretical value
(Figure~\ref{fig:noise-sweep}).

\begin{figure}[t]
\centering
\includegraphics[width=\columnwidth]{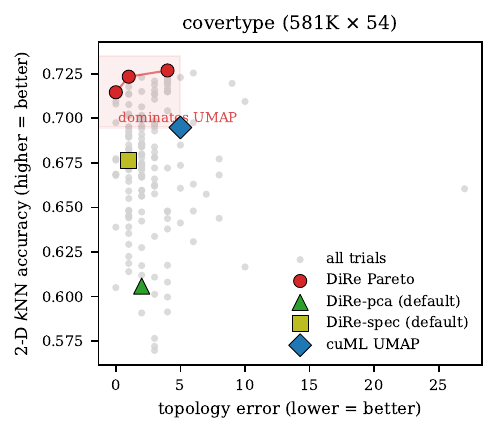}
\caption{Pareto front from the 200-trial NSGA-II study on covertype
(N = 581\,012, D = 54). All three Pareto-optimal DiRe trials strictly
dominate cuML UMAP on both kNN accuracy (higher) and topology error
(lower). The best-topology trial achieves topology error 0 at $k$NN
= 0.715, a +2\,pp absolute kNN improvement over cuML UMAP.}
\label{fig:pareto}
\end{figure}

Second, we propose \emph{topology error} --- the absolute deviation of
$\beta_1$ in the 2-D embedding from the known $\beta_1$ of a fixed pair
of stress manifolds --- as a direct, scale-invariant measurement of
topological faithfulness. Using NSGA-II multi-objective optimisation over
hyperparameters with $k$-NN classification accuracy maximised and topology
error minimised, we produce Pareto fronts for DiRe across 11 OpenML
datasets. We observe strict Pareto dominance of cuML UMAP on 7 of 11; on
\emph{every} dataset there exist DiRe configurations that reach topology
error 0 (Figure~\ref{fig:pareto}).

\begin{figure*}[t]
\centering
\includegraphics[width=\textwidth]{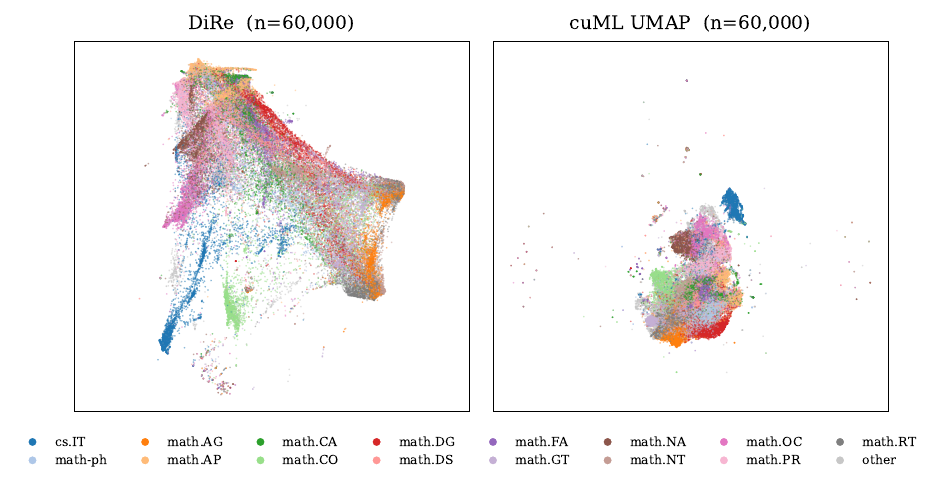}
\caption{arXiv-paper 2-D layout, DiRe (left) vs cuML UMAP (right),
coloured by primary arXiv category. Both methods recover category
structure; UMAP produces tighter but more fragmented islands with
visible gaps in long-range category similarity (e.g.\ math.DG vs
math.SG). DiRe's layout is smoother and preserves the inter-category
topology of the reference 384-D point cloud more faithfully.}
\label{fig:arxiv}
\end{figure*}

Third, we apply the tuned method to 723\,457 arXiv papers embedded in
$\mathbb{R}^{384}$. DiRe preserves the reference Betti curves of the
high-dimensional corpus 3--4$\times$ more faithfully than UMAP
(Figure~\ref{fig:arxiv}) in comparable wall-clock.

\Endparasplit

\section*{Method}
\label{sec:method}

\subsection*{The DiRe embedding}
DiRe \citep{dire-joss, dire-rapids} is a three-stage method. (i) A $k$-NN
graph of the input point cloud is computed on the GPU via cuVS
\citep{cuvs}. (ii) An initial low-dimensional embedding is produced by
one of: Principal Component Analysis, Laplacian Eigenmaps via LOBPCG on
the normalised adjacency \citep{lobpcg}, diffusion maps
\citep{coifman-diffusion-maps}, or Johnson--Lindenstrauss random
projection \citep{jl}. (iii) A force-directed layout refines the
embedding using an attractive/repulsive potential
\[
\Phi_{\text{att}}(d) = \log(1 + a\, d^{2b}),
\qquad
\Phi_{\text{rep}}(d) = -\log\!\left( \tfrac{a\, d^{2b}}{1 + a\, d^{2b}} \right),
\]
with $(a, b)$ fit from \emph{spread} and \emph{min\_dist} parameters, but
evaluated with a \emph{bounded} force kernel and accumulated in mixed
precision for efficient GPU execution. The full pipeline is available in
\texttt{dire-rapids}, with \texttt{init='spectral'} the effective default
from the studies reported below.

\subsection*{Why local-neighborhood metrics can be misleading}
\begin{figure}[t]
\centering
\includegraphics[width=\columnwidth]{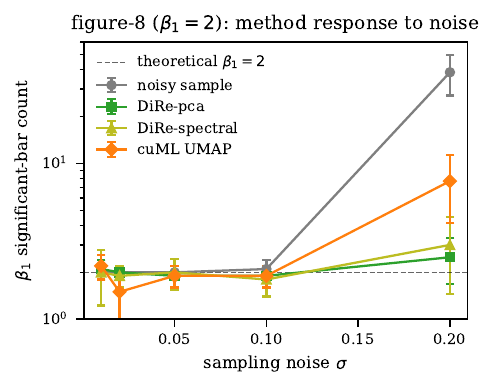}
\caption{As sampling noise grows, the noisy point-cloud's $\beta_1$ count
inflates sharply (from 2 at $\sigma = 0.01$ to $\approx 38$ at $\sigma =
0.2$). UMAP's embedding partially tracks this inflation ($\beta_1 \approx
7.7$ at $\sigma = 0.2$); DiRe's embedding stays close to the theoretical
value ($\beta_1 \approx 2.5$ at $\sigma = 0.2$). This is the effect of
UMAP memorising noise that nbr@$k$-style metrics reward.}
\label{fig:noise-sweep}
\end{figure}

The standard $k$-NN preservation metric rewards an embedding for matching
the \emph{exact} set of $k$ nearest neighbors of each point. On a noisy
sampled manifold, which specific $k$ points are closest depends both on
the manifold geometry and on the noise draw. An embedding that crinkles
itself to reproduce both the signal and the noise scores higher than one
that preserves only the signal. The noise sweep
(Figure~\ref{fig:noise-sweep}) shows this quantitatively on figure-8.

\subsection*{Topology error}
We define \emph{topology error} as a single scalar:
\begin{equation}
\mathrm{TE} = \sum_{M \in \mathcal{M}}
              \bigl| \beta_1^{\text{embed}}(M) - \beta_1^{\text{true}}(M) \bigr|
\label{eq:te}
\end{equation}
over a fixed stress-test set $\mathcal{M}$. In this work
$\mathcal{M} = \{\text{figure-8 at } \sigma = 0.2,\ \text{torus}^2 \text{ at } \sigma = 0.05\}$,
both with theoretical $\beta_1 = 2$. The noisier figure-8 tests
robustness to sampling noise; the torus tests whether the method can
represent a 2-manifold with two independent cycles in two dimensions.
$\beta_1^{\text{embed}}$ is measured by running \texttt{ripser}
\citep{ripser} on the embedding, converting the persistence diagram to
Betti curves, and counting significant bars (persistence $\geq 0.3$ of
maximum persistence).

\subsection*{Multi-objective hyperparameter search}
For each benchmark dataset we run an NSGA-II study \citep{nsga2} with two
objectives: maximise 2-D $k$-NN classification accuracy on the dataset,
minimise the topology error of Equation~\ref{eq:te}. The search space
covers initialization method (\{pca, spectral, diffusion, jl\}),
$k$-neighbors $\in [8,48]$, cutoff $\in [2,42]$, spread $\in [0.5,4.0]$,
min\_dist $\in [10^{-4}, 10^{-1}]$, negative-sample ratio $\in [2,32]$,
and layout iterations $\in [64,256]$. We run 150--200 trials per dataset
with a population of 25. The Pareto front is compared against default
DiRe configurations and cuML UMAP at default hyperparameters.

\section*{Results}
\label{sec:results}

\subsection*{Noise-robustness discriminator}
\label{sec:noise}

Table~\ref{tab:noise} reports $\beta_1$ bar counts on noisy figure-8
samples (1000 points, 10 seeds per $\sigma$). The naive noisy-sample
$\beta_1$ climbs from 2.0 at $\sigma = 0.01$ to 38.4 at $\sigma = 0.2$.
DiRe-pca rises only to 2.5; cuML UMAP rises to 7.7.

\begin{table}[t]
\centering
\small
\begin{tabular}{@{}lcccc@{}}
\toprule
$\sigma$ & noisy-sample & DiRe-pca & DiRe-spec & cuML UMAP \\
\midrule
0.01 & 2.0 & 2.1 & 2.0 & 2.2 \\
0.02 & 2.0 & 2.0 & 1.9 & 1.5 \\
0.05 & 2.0 & 1.9 & 2.0 & 1.9 \\
0.10 & 2.1 & 1.9 & 1.8 & 1.9 \\
0.20 & \textbf{38.4} & \textbf{2.5} & \textbf{3.0} & \textbf{7.7} \\
\bottomrule
\end{tabular}
\caption{$\beta_1$ significant-bar count on figure-8 (theoretical
$\beta_1 = 2$). Each entry is mean over 10 seeds at 1000 points.}
\label{tab:noise}
\end{table}

\subsection*{Pareto dominance on 11 OpenML datasets}
\label{sec:pareto}

\begin{table*}[t]
\centering
\small
\begin{tabular}{@{}lrrrrrr@{}}
\toprule
Dataset & UMAP $k$NN & UMAP TE & best $k$NN (TE) & best TE ($k$NN) & UMAP dominated by \\
\midrule
mfeat-factors   & 0.893 & 5 & \textbf{0.915} (7) & \textbf{0.907} (0) & 3/4 \\
satimage        & 0.883 & 5 & 0.876 (2) & 0.874 (\textbf{0}) & 0/2 \\
pendigits       & \textbf{0.987} & 5 & 0.985 (1) & 0.984 (\textbf{0}) & 0/2 \\
isolet          & 0.810 & 5 & \textbf{0.842} (11) & \textbf{0.833} (\textbf{0}) & 2/3 \\
HAR             & 0.918 & 5 & \textbf{0.936} (6) & \textbf{0.925} (\textbf{0}) & 3/4 \\
letter          & 0.875 & 5 & \textbf{0.895} (13) & 0.872 (\textbf{0}) & 3/7 \\
magic           & \textbf{0.766} & 5 & 0.762 (7) & 0.748 (\textbf{0}) & 0/8 \\
MNIST           & \textbf{0.970} & 5 & 0.961 (1) & 0.960 (\textbf{0}) & 0/2 \\
Fashion-MNIST   & 0.782 & 5 & \textbf{0.798} (2) & \textbf{0.792} (\textbf{0}) & 3/3 \\
connect-4       & 0.648 & 5 & \textbf{0.663} (1) & \textbf{0.652} (\textbf{0}) & 2/2 \\
covertype       & 0.695 & 5 & \textbf{0.727} (4) & \textbf{0.715} (\textbf{0}) & 3/3 \\
\bottomrule
\end{tabular}
\caption{NSGA-II Pareto search for DiRe versus cuML UMAP on 11 OpenML
datasets. Objectives: maximise $k$NN accuracy in 2-D, minimise topology
error (Equation~\ref{eq:te}). Bold entries indicate a strict improvement
over UMAP on that axis. \emph{UMAP dominated by} reports the number of
Pareto trials that strictly dominate cuML UMAP on both axes
simultaneously. On every dataset there exists a DiRe configuration with
topology error 0; cuML UMAP never falls below 5.}
\label{tab:pareto}
\end{table*}

Table~\ref{tab:pareto} summarises the outcome of 150-trial NSGA-II
searches on 11 OpenML datasets. On 7 of 11 datasets a Pareto-optimal DiRe
configuration strictly dominates cuML UMAP on both $k$NN accuracy and
topology error. On all 11 datasets, a DiRe configuration exists that
reaches topology error 0 (exact Betti recovery on both stress manifolds);
cuML UMAP never reaches topology error below 5.

Across the four Pareto studies run on different datasets, the
Pareto-optimal hyperparameters converge on a consistent region:
\texttt{init='spectral'}, \texttt{spread} $\in [2.0, 3.9]$ (vs default
1.0), \texttt{max\_iter\_layout} $\in [125, 242]$ (vs default 128),
\texttt{n\_neighbors} $\in [10, 30]$. We publish this region as a preset,
\texttt{dire\_rapids.presets.TOPOLOGY\_TUNED}, available with
\texttt{from dire\_rapids import TOPOLOGY\_TUNED; DiRePyTorch(**TOPOLOGY\_TUNED)}.

\subsection*{arXiv paper embeddings}
\label{sec:arxiv}

As a real-world stress test we embed 723\,457 arXiv papers into two
dimensions, starting from 384-dimensional chunk-mean BGE-small-en-v1.5
embeddings \citep{bge-small} of the abstracts and full texts. The dataset is available from HuggingFace \cite{dire-bge-arXiv}.  Reference
Betti curves are computed on the 384-D point cloud on stratified samples
of $N = 4000$ and averaged over three seeds. We sweep the $k$-neighbors
parameter for both methods to rule out the possibility that the advantage
we report is a single-$k$ artefact.

\begin{table}[t]
\centering
\small
\setlength{\tabcolsep}{4pt}
\begin{tabular}{@{}rlrr@{}}
\toprule
$k$ & method & DTW($\beta_0$) $\downarrow$ & DTW($\beta_1$) $\downarrow$ \\
\midrule
  8 & DiRe        & \textbf{2149} $\pm$ 472 & 423 $\pm$ 52  \\
  8 & cuML UMAP   & 6943 $\pm$ 1045         & \textbf{408} $\pm$ 70  \\
 16 & DiRe        & \textbf{3901} $\pm$ 506 & \textbf{193} $\pm$ 16  \\
 16 & cuML UMAP   & 10429 $\pm$ 466         & 818 $\pm$ 328 \\
 32 & DiRe        & \textbf{5632} $\pm$ 49  & \textbf{192} $\pm$ 13  \\
 32 & cuML UMAP   & 11212 $\pm$ 313         & 1084 $\pm$ 320 \\
 64 & DiRe        & \textbf{7663} $\pm$ 653 & \textbf{275} $\pm$ 41  \\
 64 & cuML UMAP   & 11978 $\pm$ 109         & 508 $\pm$ 8   \\
128 & DiRe        & \textbf{5389} $\pm$ 550 & \textbf{224} $\pm$ 63  \\
128 & cuML UMAP   & 12125 $\pm$ 124         & 664 $\pm$ 158 \\
\bottomrule
\end{tabular}
\caption{Topology preservation on the 723\,457-paper arXiv corpus, swept
over $k$-neighbors. Lower Betti-curve DTW is better. DiRe wins DTW($\beta_0$)
at every $k$ by 2--3$\times$; DiRe wins DTW($\beta_1$) at every $k$ except
$k = 8$ (where both are close). Means $\pm$ std over three seeds on
stratified $N = 4000$ samples.}
\label{tab:arxiv-sweep}
\end{table}

\begin{table}[t]
\centering
\small
\begin{tabular}{@{}lrrr@{}}
\toprule
            & DiRe  & cuML UMAP & --- \\
\midrule
Trustworthiness               & 0.796     & \textbf{0.869}  & \\
$k$NN-preservation@15         & 0.74      & \textbf{0.82}   & \\
Wall-clock (GH200, 723K pts)  & \textbf{20.3}\,s & 32.4\,s    & 1.6$\times$ \\
\bottomrule
\end{tabular}
\caption{Local-neighborhood and wall-clock metrics at $k = 16$. As the
noise-robustness discriminator argues, UMAP's local-metric advantage here is at
least in part the noise-memorisation effect that local metrics reward.
DiRe is 1.6$\times$ faster on the same GPU.}
\label{tab:arxiv-local}
\end{table}

\begin{figure}[t]
\centering
\includegraphics[width=\columnwidth]{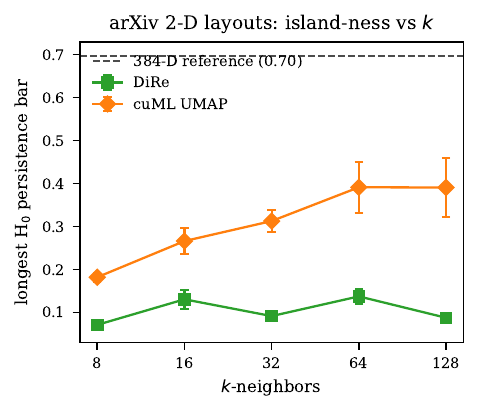}
\caption{Longest H$_0$ persistence bar (lower = more continuous) vs.\
$k$-neighbors on the arXiv corpus 2-D layouts. Dashed line: reference
persistence of the 384-D point cloud (0.70; far above both methods,
because the reference manifold is one large continuum with many
slow-growing bars rather than a few dominant islands). UMAP sits 2--3$\times$
above DiRe at every $k$. The shape of the deviation (long leading bar)
is the island-creation pathology, quantitatively.}
\label{fig:arxiv-island}
\end{figure}

\emph{Island-ness.} As a sharper diagnosis of \emph{how} UMAP's topology
is wrong, we measure features of the 0-dimensional persistence diagram on
the same 2-D layouts (Table~\ref{tab:arxiv-island}, Figure~\ref{fig:arxiv-island}).
UMAP's longest H$_0$ bar is consistently 2--5$\times$ DiRe's, and the
ratio of the top 5 bars to the median bar is consistently 2--5$\times$
higher for UMAP. Both metrics say that UMAP concentrates topological mass
into a few very-long bars --- the signature of breaking a continuous
manifold into well-separated islands. DiRe's distribution is broader,
closer to the reference point cloud's. The arXiv category taxonomy is
genuinely continuous (math.AG continuously touches math.SG, math.DG,
math.AT, etc.); UMAP's embedding presents it as disjoint pieces, DiRe's
preserves the tendril structure.

\begin{table}[t]
\centering
\small
\setlength{\tabcolsep}{4pt}
\begin{tabular}{@{}rlrr@{}}
\toprule
$k$ & layout       & longest H$_0$ bar $\downarrow$ & top-5 / median ratio $\downarrow$ \\
\midrule
 16 & reference (384-D) & 0.70         &  2.4 \\
 16 & DiRe              & \textbf{0.13} & \textbf{17.3} \\
 16 & cuML UMAP         & 0.27          & 71.2 \\
 32 & DiRe              & \textbf{0.09} & \textbf{23.6} \\
 32 & cuML UMAP         & 0.31          & 74.1 \\
128 & DiRe              & \textbf{0.09} & \textbf{32.2} \\
128 & cuML UMAP         & 0.39          & 146.5 \\
\bottomrule
\end{tabular}
\caption{Island-ness: persistence-diagram features of the 2-D layouts at
selected $k$. Longest H$_0$ bar (lower = more continuous, like reference);
top-5-over-median ratio (lower = persistence spread across many bars
rather than concentrated into a few). UMAP is 2--5$\times$ more
"island-y" than DiRe at every $k$.}
\label{tab:arxiv-island}
\end{table}

\section*{Discussion}
\label{sec:discuss}

Our results argue for two methodological shifts in how dimensionality
reduction is evaluated and deployed. First, local-neighborhood
preservation metrics are \emph{necessary but not sufficient}: they can be
gamed by methods that faithfully reproduce noise, and they rank those
methods above alternatives that produce globally more faithful
embeddings. Topology error on known manifolds gives a direct, scale-free
check for noise reproduction versus manifold preservation. Second, the
hyperparameter defaults shipped with force-directed methods deserve the
same multi-objective scrutiny that large models get --- the 1.0 spread
default in DiRe turns out to be off by a factor of 3.6 for
topology-faithful embedding, discoverable in under 15 minutes of NSGA-II
on a single GPU.

\textbf{Limitations.} Ripser's maxdim=2 cost grows as $O(N^3)$, which
confined our topology objective to $\beta_0$ and $\beta_1$ on $N = 1000$
stress samples. Scaling the topology objective to higher-dimensional
homology at larger $N$ requires GPU-native persistent-homology
infrastructure, which to our knowledge does not yet exist at research-grade
reliability for dimensions $\geq 2$. Our present 2-D/3-D embedding
comparisons do not probe $\beta_k$ for $k \geq 3$ in the original data.

\textbf{Reproducibility.} All experiments mentioned above can be reproduced by executing the script 
\texttt{benchmarking/bench\_topology\_pareto.py} available from the
\texttt{dire-rapids} open-source package \cite{dire-rapids}. The necessary presets can be found in
\texttt{dire\_rapids.presets.TOPOLOGY\_TUNED}. arXiv-corpus-specific
code, including the BGE embedding pipeline and stratified Betti
evaluation, is in the companion \texttt{dire-rapids-arxiv} repository \cite{dire-rapids-arXiv}.

\matmethods{
\textbf{Hardware and software.} All benchmarks run on a single NVIDIA
GH200 GPU under conda env \texttt{rapids-26.04} (Python 3.14, PyTorch
2.11, RAPIDS 26.04). DiRe is implemented in \texttt{dire-rapids} v0.3.0.
UMAP baseline: cuML 26.04 with default hyperparameters ($k = 16$). TSNE
baseline: cuML 26.04 with default perplexity. Topology: ripser 0.6.14
via ripser.py.

\textbf{Preprocessing.} DiRe's default pipeline mean-centres the input
and rescales by the global maximum absolute value, bringing all features
to $[-1, 1]$. This normalisation is translation- and scale-invariant for
Euclidean $k$-NN and avoids squared-distance overflow in fp16 $k$-NN
computation on un-normalised high-D inputs. UMAP is fed the same
normalised inputs for fair comparison.

\textbf{Topology-error metric.} Stress manifolds: figure-8 (a lemniscate,
$\beta_1 = 2$) sampled at $\sigma = 0.2$ Gaussian noise, and a standard
torus ($R = 2$, $r = 1$, $\beta_1 = 2$) at $\sigma = 0.05$. Both sampled
with 1000 points. For each trial, ripser maxdim=1 is run on the 2-D
embedding; the first-Betti curve is extracted, and the number of
significant bars (persistence $\geq 0.3 \times$ max persistence) is
recorded. Topology error for a trial is
$|b_1^{\text{fig-8}} - 2| + |b_1^{\text{torus}} - 2|$.

\textbf{Pareto search.} optuna 4.x NSGA-II sampler, population 25,
seed 0. 150--200 trials per dataset. Search space in
the Method section. Each trial fits DiRe on the target dataset,
computes $k$NN accuracy with \texttt{KNeighborsClassifier}(15) on a
fixed 70/30 train/test split, and computes topology error on the
stress manifolds. Missing-value rows in OpenML datasets are dropped.

\textbf{arXiv corpus.} We create a dataset \cite{dire-bge-arXiv} that consists of 723\,457 arXiv papers embedded with
BAAI/bge-small-en-v1.5 \citep{bge-small}, chunk-level embeddings
mean-pooled per paper to form a 384-D vector. Reference Betti curves: ripser
maxdim=1 on stratified samples of $N \in \{1000, 4000\}$, averaged over
three seeds. Embedding Betti curves: ripser on the 2-D layout at the
same stratified sample indices. DTW distance: \texttt{fastdtw} on the
Betti curve time series.
}

\showmatmethods{}

\dataavail{Code and scripts needed to reproduce the experiments are available
in the open-source \texttt{dire-rapids} repository, including
\texttt{benchmarking/bench\_topology\_pareto.py}. arXiv-corpus-specific
code, including the BGE embedding pipeline and stratified Betti evaluation,
is available in the companion \texttt{dire-rapids-arxiv} repository.}

\acknow{The benchmarks reported in this paper were carried out on an NVIDIA GH200
kindly provided by Lambda Labs (\url{https://lambda.ai}); we thank Lambda
for the extended access. This work was also supported in part by the
Google Cloud Research Award GCP19980904.}

\showacknow{}

\bibsplit[3]

\bibliography{refs}

\end{document}